\documentclass[10pt,letterpaper,twoside]{article}

\usepackage[T1]{fontenc}
\usepackage[english]{babel}  
\usepackage{amsmath}   
\usepackage{titlesec}  
\usepackage{enumitem}  
\usepackage{listings}  
\usepackage{graphicx}
\usepackage{float}     
\usepackage{microtype}  
\usepackage{setspace}   
\usepackage[authoryear, round]{natbib}  

\titlespacing*{\section}{0pt}{1.2ex}{1ex}  
\titlespacing*{\subsection}{0pt}{1.2ex}{1ex}
\titlespacing*{\subsubsection}{0pt}{1ex}{0.8ex}

\setkeys{Gin}{width=0.85\textwidth}    
\title{QIXAI: A Quantum-Inspired Framework for Enhancing Classical and Quantum Model Transparency and Understanding}
\author{
	John M. Willis, Sustainable Future Tech \\ 
	\texttt{QIXAI@SustainableFutureTech.com}  
}
\date{October 21, 2024}  

\begin{document}
	\maketitle
	\begin{abstract}
		Despite the impressive performance of deep learning models, particularly Convolutional Neural Networks (CNNs), their lack of interpretability and explainability renders them "black boxes." This opacity is problematic, especially in critical domains such as healthcare, finance, and autonomous systems, where trust and accountability are essential. This paper introduces the QIXAI Framework (Quantum-Inspired Explainable AI), a novel approach for enhancing neural network interpretability using layer-wise quantum-inspired methods. By leveraging principles from quantum mechanics—such as Hilbert spaces, superposition, entanglement, and eigenvalue decomposition—the QIXAI framework uncovers how different layers in neural networks process and combine features to make decisions.
		
		We critically examine popular model-agnostic interpretability methods like SHAP and LIME and layer-specific techniques like Layer-wise Relevance Propagation (LRP), identifying their limitations in providing a comprehensive understanding of neural networks' inner workings. The QIXAI framework addresses these limitations, offering deeper insights into feature importance, dependencies between layers, and information propagation across layers. Using a CNN for malaria parasite detection as a case study, we demonstrate how quantum-inspired methods such as Singular Value Decomposition (SVD), Principal Component Analysis (PCA), and Mutual Information (MI) yield actionable, interpretable explanations of model behavior. Additionally, we explore how the QIXAI framework may be extended to other architectures, such as Recurrent Neural Networks (RNNs), Long Short-Term Memory (LSTM) networks, Transformers, and Natural Language Processing (NLP) models, as well as applied across diverse domains including generative models and time-series analysis. The framework’s applicability spans both quantum and classical implementations, demonstrating its potential to improve interpretability and transparency across a wide range of models. These insights contribute to the broader goal of building transparent, trustworthy, and interpretable AI systems.
	\end{abstract}
	\section*{Introduction}
	Deep learning models, particularly Convolutional Neural Networks (CNNs), have achieved state-of-the-art results in numerous fields, such as image recognition \citep{krizhevsky2012imagenet}, medical diagnostics \citep{esteva2017dermatologist}, and natural language processing \citep{vaswani2017attention}. Despite these successes, the inner workings of these models remain largely opaque, leading to concerns about their interpretability and explainability \citep{lipton2018mythos}. In high-stakes domains like healthcare, finance, and autonomous systems, this lack of transparency can hinder trust and adoption \citep{rudin2019stop, doshi2017towards}. Consequently, interpretability has emerged as a critical need for making AI systems more accountable and transparent \citep{samek2019explainable}.
	
	Numerous methods have been developed to address this challenge, including model-agnostic techniques like SHAP \citep{lundberg2017unified} and LIME \citep{ribeiro2016should}, and layer-specific approaches like Layer-wise Relevance Propagation (LRP) \citep{montavon2017}. However, these methods often fall short of providing a comprehensive, global understanding of how different layers in a neural network contribute to predictions. Model-agnostic methods are limited in scope, offering only local interpretability for individual predictions, while LRP suffers from instability in deeper layers and lacks a rigorous mathematical framework for explaining the propagation of information across layers.
	
	In this paper, we introduce the QIXAI Framework (Quantum-Inspired Explainable AI), a novel approach for enhancing neural network interpretability through quantum-inspired mathematical methods. Drawing on concepts from quantum mechanics—such as Hilbert spaces, superposition, and entanglement—we propose a layer-wise analysis that provides more comprehensive and mathematically grounded insights into neural network behavior. These quantum-inspired methods, including Singular Value Decomposition (SVD), Principal Component Analysis (PCA), and Mutual Information (MI), offer a deeper understanding of feature importance, layer correlations, and how information flows through the network.
	
	As a case study, we apply these techniques to a classical CNN developed for malaria parasite detection. We demonstrate that quantum-inspired methods offer significant advantages over traditional interpretability approaches, revealing how features are extracted and combined at different layers. Furthermore, we explore the applicability of the QIXAI Framework to other architectures, such as Recurrent Neural Networks (RNNs), Long Short-Term Memory (LSTM) networks, Transformers, and Natural Language Processing (NLP) models. We also discuss the potential for applying this framework to generative models, time-series analysis, and more.
	
	By bridging quantum theory with classical neural networks, this work aims to make AI systems more interpretable, transparent, and trustworthy across a wide range of applications.
	\section*{Quantum-Inspired Methods for Interpretability}
	In this section, we consolidate the quantum-inspired mathematical concepts and methods used to enhance the interpretability of CNNs. Each method is presented with its theoretical foundation, implementation details, relevant outputs, and the significance of the results obtained through their application to the malaria detection CNN.
	\section{Quantum-Inspired Methods for Neural Network Analysis}
	In both Hilbert Spaces and Superposition analyses, we apply similar methods for extracting and analyzing feature activations from the neural network. Specifically, we compute cosine similarity between pooled activations from Conv1, Conv2, and Dense layers to quantify feature alignment and combination. The following steps outline the common methodology used across both sections.
	\subsection{Methodology}
	First, activations are extracted from Conv1, Conv2, and Dense layers for all samples in the test set. Global Average Pooling (GAP) is applied to reduce dimensionality, followed by Principal Component Analysis (PCA) to further reduce the feature space for more computationally efficient cosine similarity calculations. Cosine similarity is then computed between the activations of the respective layers.
	\begin{lstlisting}
		conv1_pooled = tf.reduce_mean(conv1_activations, axis=[1, 2])
		conv2_pooled = tf.reduce_mean(conv2_activations, axis=[1, 2])
		dense_activations = dense_layer_activations  # No pooling needed for dense layer
		
		# Apply PCA to reduce dimensionality
		pca = PCA(n_components=32)
		conv1_reduced = pca.fit_transform(conv1_pooled.numpy())
		conv2_reduced = pca.fit_transform(conv2_pooled.numpy()[:, :32])
		dense_reduced = pca.fit_transform(dense_activations.numpy()[:, :32])
		
		# Compute cosine similarity between the reduced activations
		similarity_conv1_conv2 = cosine_similarity(conv1_reduced, conv2_reduced)
		similarity_conv1_dense = cosine_similarity(conv1_reduced, dense_reduced)
		similarity_conv2_dense = cosine_similarity(conv2_reduced, dense_reduced)
	\end{lstlisting}
	\subsection{Results}
\begin{figure}[h!]
		\centering
		\includegraphics[width=0.8\textwidth]{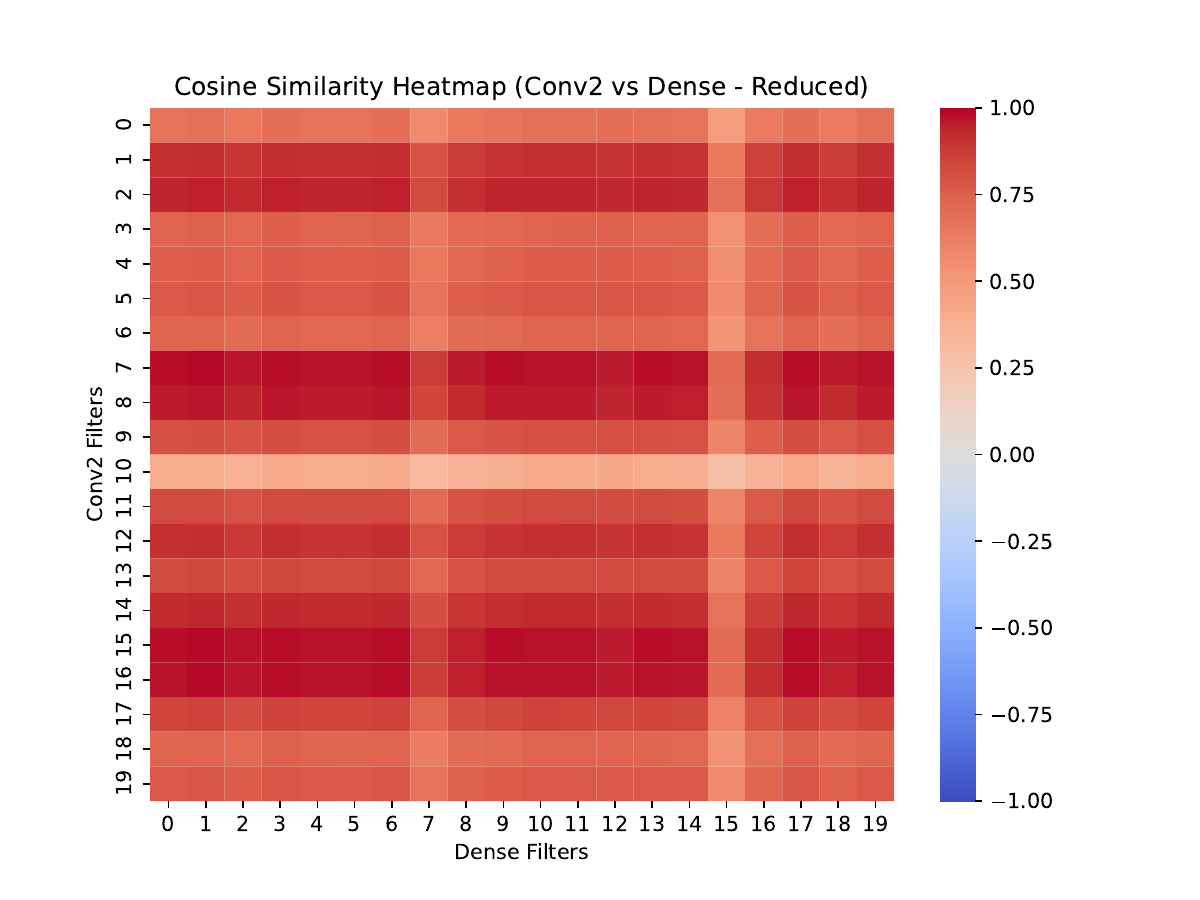}
		\caption{Cosine Similarity Summary Heatmap (Conv2 vs Dense Layer).}
		\label{fig:cosine_heatmap}
	\end{figure}

	The cosine similarity metrics between Conv1, Conv2, and Dense layer activations are summarized in Table~\ref{tab:cosine_similarity_1}.
	\begin{table}[h!]
		\centering
		\begin{tabular}{|c|c|}
			\hline
			\textbf{Layer Comparison} & \textbf{Cosine Similarity} \\ \hline
			Conv1 vs Conv2 & 0.0032 \\ \hline
			Conv1 vs Dense & -0.0195 \\ \hline
			Conv2 vs Dense & 0.0194 \\ \hline
		\end{tabular}
		\caption{Cosine similarity between Conv1, Conv2, and Dense layer activations.}
		\label{tab:cosine_similarity_1}
	\end{table}
	A summary heatmap comparing the cosine similarity between Conv2 and Dense layer filters is shown in Figure~\ref{fig:cosine_heatmap}.
	\section{Hilbert Spaces and Vector Representations}
	\subsection{Concept and Theoretical Foundation}
	Hilbert spaces are fundamental in quantum mechanics for representing quantum states as vectors in high-dimensional spaces \citep{nielsen2000}. In a neural network, the activations from different layers can also be viewed as vectors in such spaces, where vector operations like the inner product and cosine similarity provide insights into how activations from different layers relate to each other.
	
	The inner product measures the degree to which two vectors are aligned. For activations from layers such as Conv1 and Conv2, the inner product \( \langle a_{\text{Conv1}}, a_{\text{Conv2}} \rangle \) quantifies the overlap between feature maps from the two layers. Mathematically, it is expressed as:
	\[
	\langle a_{\text{Conv1}}, a_{\text{Conv2}} \rangle = \sum_{i=1}^{n} a_{\text{Conv1}}^{(i)} \cdot a_{\text{Conv2}}^{(i)}
	\]
	where \( a_{\text{Conv1}}^{(i)} \) and \( a_{\text{Conv2}}^{(i)} \) are feature maps from Conv1 and Conv2, respectively.
	
	While the inner product gives a raw measure of alignment, it does not account for the magnitudes of the vectors, which can make it difficult to interpret. Therefore, we use cosine similarity to normalize the vectors and better understand their angular relationship. Cosine similarity is defined as:
	\[
	\cos \theta = \frac{\langle a_{\text{Conv1}}, a_{\text{Conv2}} \rangle}{\|a_{\text{Conv1}}\| \|a_{\text{Conv2}}\|}
	\]
	where \( \cos \theta \) measures the angle between the vectors. A cosine similarity near zero suggests orthogonality, meaning that the layers capture nearly independent features.
	\subsection{Results and Interpretation}
	The results in Table 2 were obtained from the analysis of the Conv1 and Conv2 activations.
	\begin{table}[h!]
		\centering
		\begin{tabular}{|c|c|}
			\hline
			\textbf{Metric} & \textbf{Value} \\ \hline
			Conv1 Inner Product & -0.0024 \\ \hline
			Conv1-Conv2 Cosine Similarity & 0.0032 \\ \hline
		\end{tabular}
		\caption{Inner product and cosine similarity between Conv1 and Conv2 activations.}
		\label{tab:cosine_similarity_2}
	\end{table}
	The inner product of \(-0.0024\) indicates a very small degree of overlap between the features learned by Conv1 and Conv2. When normalized to compute cosine similarity, we find a value of \( 0.0032 \), suggesting that the two layers learn almost orthogonal (independent) features. This orthogonality aligns with the hierarchical nature of CNNs, where early layers like Conv1 capture basic patterns, while deeper layers like Conv2 focus on more abstract patterns \citep{manning2008introduction}.
	\section{Superposition and Feature Combination}
	\subsection{Concept and Theoretical Foundation}
	Superposition in quantum mechanics refers to the ability of a system to exist simultaneously in multiple states \citep{schuld2018}. In neural networks, this concept translates to the combination of feature representations from different layers, where features from Conv1, Conv2, and Dense layers are combined to form the final prediction.
	
	Cosine similarity between layers helps quantify how well features are combined:
	\[
	a_{\text{final}} = w_1 a_{\text{Conv1}} + w_2 a_{\text{Conv2}} + w_3 a_{\text{Dense}}
	\]
	where \( w_1, w_2, w_3 \) are learned weights that determine the contribution of each layer to the final prediction.
	\subsection{Implementation}
	Cosine similarity was computed between activations from Conv1, Conv2, and the Dense Layer to analyze the relationships between the features extracted by these layers. The activations were pooled using Global Average Pooling (GAP) to reduce dimensionality, and Principal Component Analysis (PCA) was applied to further reduce the feature space. Cosine similarity was then calculated between Conv1 and Conv2, Conv1 and Dense, and Conv2 and Dense layers. This method reveals which layers capture correlated, redundant, or independent information, highlighting how features are combined across layers.
	
	A summary heatmap comparing the cosine similarity between Conv2 and Dense layer filters is shown in Figure~\ref{fig:cosine_heatmap}.
	
\begin{figure}[h]
		\centering
		\includegraphics[width=0.8\textwidth]{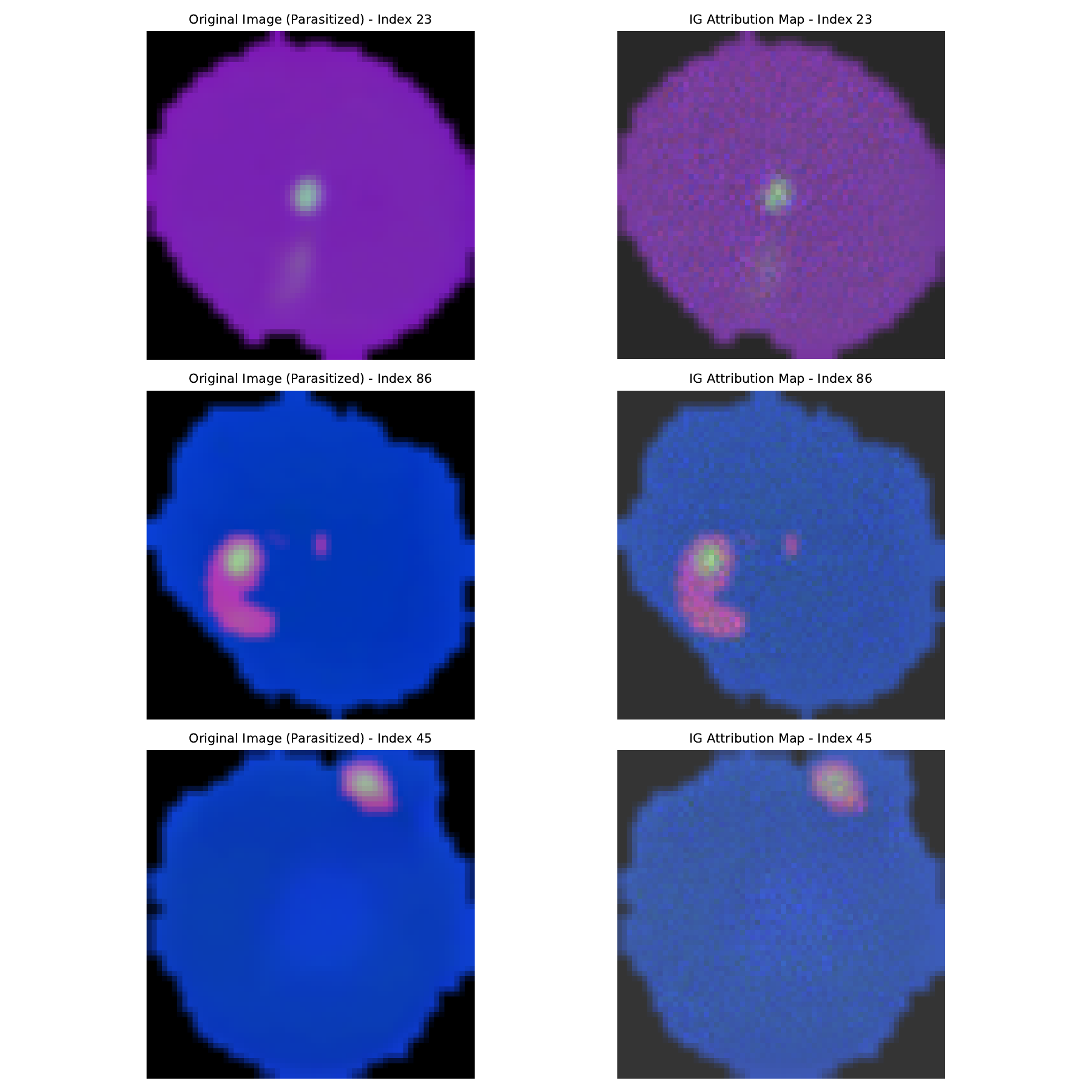}
		\caption{Integrated Gradients Attribution Maps alongside original enhanced parasitized images, showing the regions most influential in the model’s prediction.}
		\label{fig:IG_Attribution_Maps}
	\end{figure}

	In addition to the cosine similarity analysis, we employed Integrated Gradients (IG) to visualize the contributions of different pixels in the input images toward the model's decision. These attributions highlight how the network combines and processes input features across layers. The visualization of these IG attributions for three randomly selected parasitized images is presented in Figure~\ref{fig:IG_Attribution_Maps}, where both the original images and their corresponding IG attribution maps are shown. These maps offer insight into the specific areas of the input images that most strongly influence the model’s predictions.
	\subsection{Results}
	The computed cosine similarity values between the layers are summarized in Table~\ref{tab:cosine_similarity_3}.
	\begin{table}[h]
		\centering
		\begin{tabular}{|c|c|}
			\hline
			\textbf{Layer Comparison} & \textbf{Cosine Similarity Value} \\ \hline
			Conv1 vs Conv2 & 0.0032 \\ \hline
			Conv1 vs Dense & -0.0195 \\ \hline
			Conv2 vs Dense & 0.0194 \\ \hline
		\end{tabular}
		\caption{Cosine similarity values between Conv1, Conv2, and Dense layer activations.}
		\label{tab:cosine_similarity_3}
	\end{table}
	\subsection{Interpretation}
	The slightly higher similarity between Conv2 and Dense layer (0.0194) indicates that Conv2 retains more relevant information for decision-making. This suggests that features from Conv2 are more aligned with those from the Dense layer, reflecting a gradual progression of feature combination.
	
	The Integrated Gradients attribution maps (Figure~\ref{fig:IG_Attribution_Maps}) provide further support for this interpretation. The attribution maps show that specific regions in the parasitized images contribute more significantly to the model's predictions, aligning with the superposition of relevant feature representations across different layers. This combination of cosine similarity analysis and visual attributions offers a more comprehensive understanding of how the neural network processes and combines features to make its final decision.
	\section{Entanglement and Feature Correlation}
	\subsection{Concept and Theoretical Foundation}
	In quantum mechanics, entanglement refers to the strong correlation between quantum systems such that their states cannot be described independently \citep{preskill2018}. Similarly, in neural networks, mutual information (MI) measures the amount of shared information between two feature maps, quantifying their dependence. 
	
	Mathematically, mutual information \( I(X; Y) \) between two variables \( X \) and \( Y \) (e.g., feature maps from Conv1 and Conv2) is given by:
	\[
	I(X; Y) = \sum_{x \in X} \sum_{y \in Y} p(x, y) \log \left( \frac{p(x, y)}{p(x)p(y)} \right)
	\]
	where \( p(x, y) \) is the joint probability distribution of Conv1 and Conv2 activations, and \( p(x) \) and \( p(y) \) are the marginal probabilities. High MI values indicate strong correlations, akin to entanglement in quantum theory.
	\subsection{Implementation}
	To quantify dependencies between layers, mutual information (MI) was computed between the pooled activations of Conv1 and Conv2, as well as between specific feature maps.
	\begin{lstlisting}
		from sklearn.metrics import mutual_info_score
		mi = mutual_info_score(np.digitize(conv1_pooled.flatten(), bins=20),
		np.digitize(conv2_pooled.flatten(), bins=20))
	\end{lstlisting}
	\subsection{Results}
	The mutual information analysis revealed the following:
	\begin{itemize}
		\item \textbf{Mutual Information (Conv1 vs Conv2)}: 0.0016
		\item \textbf{Top MI between Feature Maps (Conv1 Map 22 vs Conv2 Map 46)}: 0.7208
		\item \textbf{Additional MI (Conv1 Map 16 vs Conv2 Map 38)}: 0.6929
	\end{itemize}
	The low overall MI score (0.0016) suggests that Conv1 and Conv2 layers extract largely independent features. However, specific feature maps show high MI, indicating strong interdependencies analogous to quantum entanglement.
	\section{Quantum Measurement and Feature-Level Interpretability}
	\subsection{Concept and Theoretical Foundation}
	In quantum mechanics, measurement collapses a superposition of states into a single observable outcome \citep{nielsen2000}. Similarly, in neural networks, methods like Integrated Gradients (IG) quantify the contribution of each input feature to the model's prediction, providing interpretability at the feature level. This is akin to quantum measurement, where certain aspects of the model collapse into interpretable insights based on gradients.
	
	Integrated Gradients are computed as:
	\[
	IG(x) = (x - x') \times \int_{\alpha=0}^{1} \frac{\partial f(x' + \alpha(x - x'))}{\partial x} d\alpha
	\]
	where \( f(x) \) is the model’s output, \( x \) is the input, and \( x' \) is a baseline input.
	\subsection{Implementation}
	Integrated Gradients were used to compute attributions for individual input features. In parallel, PCA was applied to analyze how variance is distributed across the components of the Dense layer activations.
	\begin{lstlisting}
		pca.fit(dense_activations)
		attributions = integrated_gradients_fixed(logit_model, sample_input, baseline, steps=100, sub_batch_size=10)
	\end{lstlisting}
	\subsection{Results}
	\begin{itemize}
		\item \textbf{Mean Attribution}: -0.0057
		\item \textbf{Top Positive Attribution}: 0.6526
		\item \textbf{Explained Variance by Top 5 PCA Components}:  
		\begin{itemize}
			\item Component 1: 36.96\%
			\item Component 2: 10.56\%
			\item Component 3: 7.69\%
		\end{itemize}
	\end{itemize}
	\subsection{Interpretation}
	The Integrated Gradients results highlight which parts of the input image contribute most to the model’s predictions. The top PCA component explains 36.96\% of the variance in Dense layer activations, suggesting that the network relies on a small number of key features. This reflects the collapse of a complex superposition (input) into dominant, interpretable features, analogous to quantum measurement.
	\section{Eigenvalue Decomposition for Feature Importance}
	\subsection{Concept and Theoretical Foundation}
	In quantum mechanics, eigenvalue decomposition reveals the measurable properties of a system. Similarly, Singular Value Decomposition (SVD) and Principal Component Analysis (PCA) decompose neural network activations into their most significant components. 
	
	SVD is expressed as:
	\[
	W = U \Sigma V^T
	\]
	where \( U \) and \( V \) are orthogonal matrices, and \( \Sigma \) is a diagonal matrix of singular values. The largest singular values correspond to the most significant features extracted by the network.
	\subsection{Implementation}
	SVD and PCA were applied to the Dense layer activations to identify which components carried the most significant information.
	\begin{lstlisting}
		U, S, Vt = np.linalg.svd(dense_activations, full_matrices=False)
		variance_explained = np.cumsum(S) / np.sum(S)
		pca_result, explained_variance = perform_pca(dense_activations, n_components=10)
	\end{lstlisting}
	\subsection{Results}
\begin{figure}[h!]
		\centering
		\includegraphics[width=0.8\textwidth]{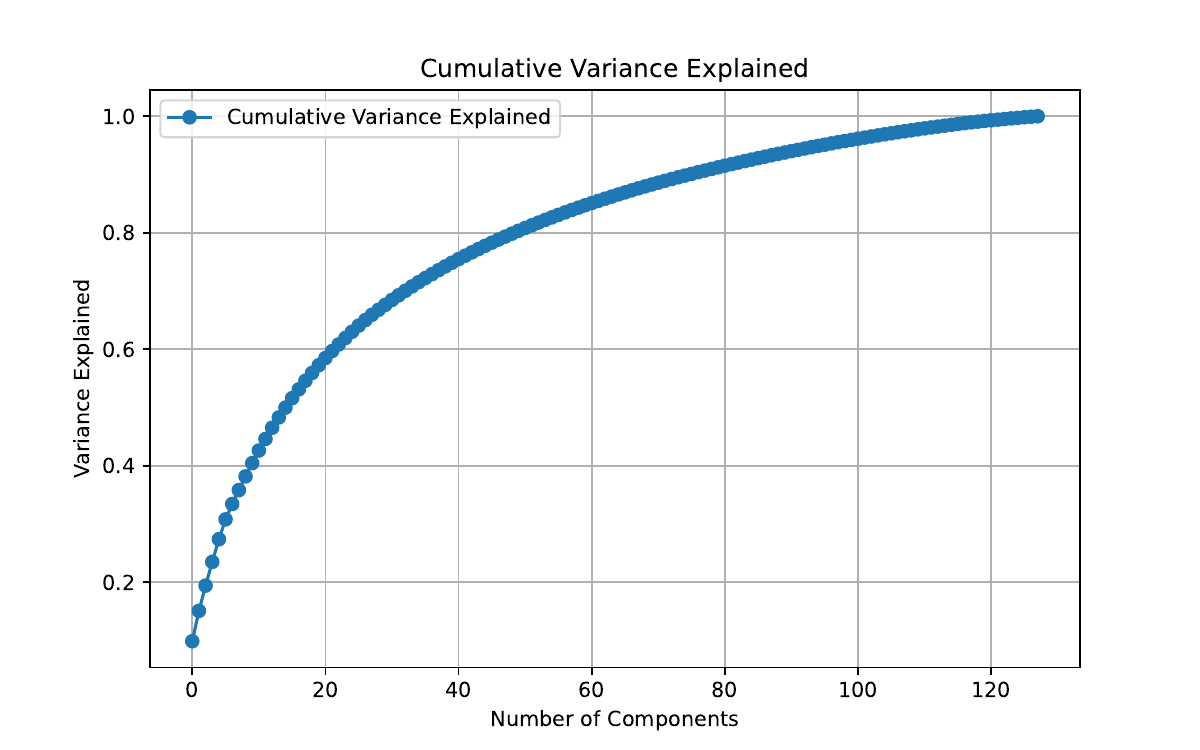}
		\caption{Cumulative Variance Explained by the Top Principal Components.}
		\label{fig:cumulative_variance}
	\end{figure}

	The cumulative variance explained by the top five components is visualized in Figure~\ref{fig:cumulative_variance}, and Table 4 below summarizes the variance explained by each of the top five components.
	\begin{table}[h!]
		\centering
		\begin{tabular}{|c|c|}
			\hline
			\textbf{Principal Component} & \textbf{Variance Explained} \\ \hline
			Component 1 & 9.88\% \\ \hline
			Component 2 & 15.10\% \\ \hline
			Component 3 & 19.43\% \\ \hline
			Component 4 & 23.51\% \\ \hline
			Component 5 & 27.40\% \\ \hline
		\end{tabular}
		\caption{Variance Explained by Top 5 Principal Components.}
		\label{tab:variance_explained}
	\end{table}
	\subsection{Interpretation}
	SVD and PCA allow us to isolate the most important features in the Dense Layer activations. The largest singular values indicate key patterns that drive the model's decisions, analogous to dominant eigenvalues in quantum systems. The cumulative variance explained in Figure~\ref{fig:cumulative_variance} shows that the first five principal components capture approximately 27.40\% of the total variance in the Dense layer activations. This suggests that a small number of components dominate the network's behavior, providing valuable insight into feature importance.
	\section*{Discussion}
	\subsection*{Key Insights}
	This study demonstrated that quantum-inspired methods, such as Integrated Gradients (IG), Singular Value Decomposition (SVD), and Mutual Information (MI), significantly enhance the interpretability of Convolutional Neural Networks (CNNs) by providing deeper insights into how features evolve across layers and which features are most critical for decision-making. These methods, including Hilbert spaces, superposition, and entanglement, were applied effectively to analyze feature dependencies, measure feature combinations, and optimize decision-making processes.
	
	Beyond CNNs, these quantum-inspired techniques may also be applied in the classical realm, extending their benefits to a wide range of machine learning architectures such as Recurrent Neural Networks (RNNs), Long Short-Term Memory (LSTM) networks, Transformers, Generative Adversarial Networks (GANs), Autoencoders, Support Vector Machines (SVMs), and Decision Trees. The QIXAI framework may provide interpretability and computational efficiency in both quantum and classical implementations of these models. In each case, quantum-inspired methods offer unique advantages for analyzing feature interactions, measuring dependencies, and isolating key components, demonstrating their versatility and potential to improve transparency across various machine learning paradigms.
	\subsection*{Limitations}
	Although quantum-inspired methods within the QIXAI framework offer valuable insights into model interpretability and decision-making processes, they can introduce additional computational overhead during the inference stage, particularly when analyzing large models or datasets. These complexities are generally associated with the quantum-inspired analysis techniques applied post-training, and not with the need to retrain the models. While the QIXAI framework is effective for enhancing interpretability across various models—such as Convolutional Neural Networks (CNNs) and others discussed later in this paper—certain methods may be better suited to models with more regular structure, like convolutional or recurrent networks. Models such as Transformers and Autoencoders may require further adaptation to fully benefit from these quantum-inspired methods.
	
	Despite these challenges, the framework's post-training, inference-focused application allows for flexibility in its integration with existing models. The following sections explore how the QIXAI framework could be further optimized and extended to other machine learning architectures, offering pathways to address these limitations and enhance performance across diverse models.
	
	\section*{Future Work and Broader Applications of the QIXAI Framework}
	While this work primarily focuses on enhancing CNN interpretability for malaria detection, the QIXAI framework has broader potential across both quantum and classical implementations of various machine learning models and techniques. By leveraging quantum concepts such as Hilbert spaces, superposition, entanglement, and quantum measurement, the framework offers significant improvements in both interpretability and computational efficiency. These methods may be applied not only to neural network architectures like Fully Connected Networks (FCNs), Recurrent Neural Networks (RNNs), Long Short-Term Memory (LSTM) networks, Transformers, Generative Adversarial Networks (GANs), Autoencoders, and Natural Language Processing (NLP) models, but also to classical machine learning approaches, such as Support Vector Machines (SVMs) and Decision Trees, as well as probabilistic models like Bayesian Networks. Whether through quantum or classical implementations, the QIXAI framework can enhance interpretability, optimize feature extraction, and improve decision-making across these diverse paradigms. Below, we explore how these quantum-inspired techniques could be integrated into these various models and methods.
	
	The QIXAI framework may apply to Fully Connected Networks (FCNs) by utilizing Hilbert spaces to map neuron activations, providing a mathematical foundation to analyze neuron interactions, as explored in Reproducing Kernel Hilbert Spaces (RKHS) theories \citep{Chen_2023}. Superposition could enhance understanding of how inputs combine across neurons, similar to quantum-inspired models in complex-valued neural networks \citep{lai_et_al_2023}. Entanglement, through mutual information, can help identify dependent neurons, offering insights into feature correlations, as demonstrated in quantum-inspired models of neural networks \citep{schuld2015simulating}. Finally, eigenvalue decomposition may isolate key weights, revealing the most critical connections driving decisions, as supported by eigenvalue regularization techniques in deep networks \citep{zhou2017eigenvalue, Chen_2023}.
	
	The QIXAI framework may apply to Recurrent Neural Networks (RNNs) by using Hilbert spaces to track the evolution of hidden states over time, providing a structured mathematical framework for analyzing temporal dependencies \citep{arXiv_1803_07128}. Superposition could be leveraged to understand how features combine across time steps, enhancing interpretability of sequential data \citep{lai_et_al_2023}. Entanglement may be used to measure dependencies between different time steps, helping to identify important correlations, as demonstrated in quantum LSTM models \citep{cai2018recurrent, arXiv_1803_07128}, while quantum measurement could highlight the most significant time steps that contribute to model predictions. Eigenvalue decomposition might pinpoint key state transitions that play a critical role in the model’s decision-making process, as explored in eigenvalue-based analysis of recurrent networks \citep{mhammedi2017efficient, Chen_2023}.
	
	In Long Short-Term Memory (LSTM) networks, the QIXAI framework may apply by using Hilbert spaces to analyze gate activations, offering a mathematical foundation for understanding interactions between the forget, input, and output gates. Superposition could explain how temporal features combine, as demonstrated in quantum LSTM models, which leverage parallelism through quantum gates like Hadamard. Entanglement may assess dependencies between gates, helping to manage long-term dependencies, as explored in quantum LSTMs leveraging entanglement for temporal correlations \citep{zhang2020quantum}, while quantum measurement could trace the impact of each gate on predictions by measuring quantum states. Finally, eigenvalue decomposition could identify key neurons involved in memory management, using quantum transforms for more efficient processing and interpretation, as highlighted in eigenvalue-based learning dynamics in deep networks \citep{saxe2014exact, arXiv_1803_07128}.
	
	In Natural Language Processing (NLP), the QIXAI framework may enhance interpretability in models like Transformers and Recurrent Neural Networks (RNNs), commonly used for tasks such as language translation and text classification. By leveraging Hilbert spaces, the framework could model relationships between words and phrases in high-dimensional space, providing insights into how linguistic features are represented across layers \citep{schuld2015simulating}. Superposition may reveal how word embeddings and contextual information combine to form sentence representations, offering a clearer view of how different layers contribute to meaning \citep{biamonte2017quantum}. Entanglement can be used to capture long-range dependencies between words, often missed by classical approaches, improving the understanding of contextual relationships \citep{liu2021rigorous}. Eigenvalue decomposition could isolate key patterns within attention mechanisms, optimizing the analysis of relationships between words, phrases, and larger text structures \citep{tacchino2019artificial}. These quantum-inspired techniques may significantly improve transparency and interpretability in NLP models.
	
	In Transformers, the QIXAI framework may apply by using Hilbert spaces to represent attention heads, enabling deeper analysis of their interactions. Superposition could explain how these heads combine, while entanglement may uncover correlations between them, as explored in quantum Transformers leveraging entanglement for attention mechanisms \citep{li2022quantum}. Quantum measurement might trace key attention weights, and eigenvalue decomposition could isolate important heads for optimization, as demonstrated in eigenvalue-based techniques for improving attention heads in Transformers \citep{yang2019breaking}.
	
	In Generative Adversarial Networks (GANs), the QIXAI framework could use Hilbert spaces to model the interaction between the generator and discriminator, providing a more structured analysis of how these components evolve in quantum state space. Superposition could reveal how generated features combine, as seen in Quantum GANs (QGANs), where quantum states offer a richer representation of data. Entanglement may measure correlations between real and generated features, highlighting dependencies that improve the model’s fidelity, particularly when leveraging entangling quantum circuits for training \citep{cincio2021entangling}. Quantum measurement could trace contributions of individual features by focusing on key quantum states during the generation process, while eigenvalue decomposition may identify critical features responsible for realistic outputs, helping optimize the network by isolating key patterns \citep{niu2022, Chang_2021}.
	
	In Autoencoders, the QIXAI framework can leverage Hilbert spaces to analyze latent space activations, offering a quantum perspective on compressed data representation. Superposition might explore how encoder-decoder features combine, aiding reconstruction, while entanglement, as demonstrated in quantum autoencoder models, can capture and measure the dependencies between latent and input features, improving the efficiency of data compression and reconstruction  \citep{romero2017}. Quantum measurement may trace the role of individual latent features in reconstruction, and eigenvalue decomposition can isolate critical latent features, improving model efficiency. Research into quantum autoencoders has shown these methods enhance compression and feature analysis \citep{romero2017, niu2022}.
	
	In Support Vector Machines (SVMs), the QIXAI framework may apply by using entanglement to measure dependencies between support vectors and the hyperplane. This could be achieved by leveraging quantum kernel methods that encode input data into high-dimensional quantum states, where entanglement measures can reveal relationships between support vectors and their influence on the decision boundary \citep{niu2022, rebentrost2014quantum}. Quantum measurement could trace the influence of these support vectors on classification decisions, offering a way to track how individual vectors contribute to the model's output \citep{suzuki2023}. Finally, eigenvalue decomposition could be used to identify critical support vectors by analyzing the quantum state corresponding to the support vectors, isolating those that play a key role in defining the hyperplane and optimizing classification accuracy \citep{romero2017, cristianini2000introduction, suzuki2023}.
	
	In Decision Trees and Random Forests, the QIXAI framework could use entanglement to measure feature dependencies across splits, enhancing feature interaction analysis. This approach has been explored in quantum machine learning models, where entanglement helps capture and optimize feature dependencies in decision-making processes \citep{gao2018quantum}. Quantum measurement may trace the importance of features in decision paths, while eigenvalue decomposition could identify critical splits by analyzing decision node eigenstates, optimizing the tree structure \citep{Farhi_1998, suzuki2023}.
	
	In Bayesian Networks, the QIXAI framework could use Hilbert spaces to model variable relationships, with superposition showing how variables combine in probabilistic inferences. Entanglement may reveal dependencies between variables, as explored in quantum Bayesian inference frameworks \citep{leifer2013towards}, while quantum measurement could trace the contributions of individual nodes. Finally, eigenvalue decomposition may highlight key variables by isolating critical nodes for accurate inferences, as demonstrated through quantum matrix factorization techniques in probabilistic graphical models \citep{larocca2019quantum}.
	\section*{Conclusion: The Importance of the QIXAI Framework for Interpretability}  
	This paper introduces the QIXAI Framework (Quantum-Inspired Explainable AI), a novel approach for enhancing neural network interpretability by leveraging quantum-inspired methods. The QIXAI framework provides a mathematically rigorous, layer-wise approach to understanding how different features are processed and combined at various stages of a neural network. By applying quantum principles such as Hilbert spaces, superposition, and entanglement, the framework uncovers feature importance, layer correlations, and the flow of information within a network.
	
	Our analysis, demonstrated through a CNN for malaria parasite detection, shows that quantum-inspired techniques like Singular Value Decomposition (SVD), Principal Component Analysis (PCA), and Mutual Information (MI) may provide clearer, more interpretable insights than traditional methods like SHAP, LIME, or Layer-wise Relevance Propagation (LRP). The QIXAI framework offers a systematic solution for overcoming the limitations of existing model-agnostic methods by providing both global and layer-wise explanations.
	
	While our primary focus has been on CNNs, the QIXAI framework may also be extended to other architectures, such as Recurrent Neural Networks (RNNs), Long Short-Term Memory (LSTM) networks, and Transformers, as well as applied to diverse domains including natural language processing and time-series forecasting. Additionally, the framework may be adapted to classical implementations of these models, broadening its impact. Future research should explore its applicability to hybrid quantum-classical systems to further enhance interpretability and computational efficiency, particularly in larger or more complex models.
	
	By establishing the QIXAI framework, we take a significant step toward creating AI systems that are not only accurate and powerful but also transparent, interpretable, and accountable—especially in high-stakes fields such as healthcare, finance, and autonomous systems.
	\section*{Acknowledgments}
	This work was supported and funded by Sustainable Future Tech.
	
	This work is licensed under a Creative Commons Attribution-NonCommercial-ShareAlike 4.0 International License (CC BY-NC-SA 4.0). To view a copy of this license, visit https://creativecommons.org/licenses/by-nc-sa/4.0/.

\end{document}